# Left Ventricle Segmentation via Optical-Flow-Net from Short-axis Cine MRI: Preserving the Temporal Coherence of Cardiac Motion

Wenjun Yan[1], Yuanyuan Wang[1(✉)], Zeju Li[1], Rob J. van der Geest[2], Qian Tao[2(✉)]

[1] Department of Electrical Engineering, Fudan University, Shanghai, China
yywang@fudan.edu.cn

[2] Department of Radiology, Leiden University Medical Center, Leiden, the Netherlands
Q.Tao@lumc.nl

**Abstract.** Quantitative assessment of left ventricle (LV) function from cine MRI has significant diagnostic and prognostic value for cardiovascular disease patients. The temporal movement of LV provides essential information on the contracting/relaxing pattern of heart, which is keenly evaluated by clinical experts in clinical practice. Inspired by the expert way of viewing Cine MRI, we propose a new CNN module that is able to incorporate the temporal information into LV segmentation from cine MRI. In the proposed CNN, the optical flow (OF) between neighboring frames is integrated and aggregated at feature level, such that temporal coherence in cardiac motion can be taken into account during segmentation. The proposed module is integrated into the U-net architecture without need of additional training. Furthermore, dilated convolution is introduced to improve the spatial accuracy of segmentation. Trained and tested on the Cardiac Atlas database, the proposed network resulted in a Dice index of 95% and an average perpendicular distance of 0.9 pixels for the middle LV contour, significantly outperforming the original U-net that processes each frame individually. Notably, the proposed method improved the temporal coherence of LV segmentation results, especially at the LV apex and base where the cardiac motion is difficult to follow.

## 1 Introduction

### 1.1 Left Ventricle Segmentation

Cardiovascular disease is a major cause of mortality and morbidity worldwide. Accurate assessment of cardiac function is very important for diagnosis and prognosis of cardiovascular disease patients. Cine magnetic resonance imaging (MRI) is the current gold standard to assess the cardiac function [1], covering different imaging planes (around 10) and cardiac phases (ranging from 20 to 40).

The large number of total images (200-400) poses significant challenges for manual analysis in clinical practice, therefore computer-aided analysis of cine MRI has been actively studied for decades. Most traditional methods in literature are based on dedicated mathematical models of shape and intensity[2]. However, the substantial variations in the cine images, including the acquisition parameters, image quality, heart morphology/pathology, etc., all make it too challenging, if not impossible, for traditional image analysis methods to reach a clinically acceptable balance of accuracy, robustness, and generalizability. As such, in current practice, the analysis of cine images still involves significant manual work, including contour tracing, or initialization and correction to aid semi-automated computer methods.

Current development of deep Convolutional Neural Networks (CNN) has made revolutionary improvement on many medical image analysis problems, including automated cine MRI analysis [3], [4]. In most of the CNN-based framework for cine MRI, nevertheless, the segmentation problem is still formulated as learning a label image from a given cine image, i.e. each frame is individually processed and there is no guarantee of temporal coherence in the segmentation results.

### 1.2 Our Motivation and Contribution

This is in contrast to what we have observed in clinical practice, as clinical experts always view the cine MRI as a temporal sequence instead of individual frames, paying close attention to the temporally-resolving motion of the heart. Inspired by the expert way of view cine MRI, we aim to integrate the temporal information to guide and regulate LV segmentation, in an easily interpretable manner.

Between temporally neighboring frames, there are two types of useful information: (1) *Difference*: the relative movement of the object between neighboring frames, providing clues of object location and motion. (2) *Similarity*: sufficient coherence exists between temporally neighboring frames, with the temporal resolution of cine set to follow cardiac motion. In this work, we proposed to use optical flow to extract the object location and motion information, while aggregating such information over a moving time window to enforce temporal coherence. Both difference and similarity measures were formulated into one module, named “optical flow feature aggregation sub-network”, which is integrated into the U-net architecture. Compared to the prevailing recurrent neural network (RNN) applied to temporal sequences [4], our method eliminates the need of introducing massive learnable RNN parameters, while preserving the simplicity and elegancy of U-net. In relatively simple scenarios like cine MRI, our proposed method has high interpretability and low computation cost.

## 2 Method

### 2.1 Optical Flow in Cine MRI

Given two neighboring temporal frames in cine MRI, the optical flow field can be calculated to infer the horizontal and vertical motion of objects in image [4], by the following equation and constraint:

$$I(x, y, t) = I(x + \Delta x, y + \Delta y, t + \Delta t) \quad (1)$$

$$\frac{\partial I}{\partial x} V_x + \frac{\partial I}{\partial y} V_y + \frac{\partial I}{\partial t} = 0 \quad (2)$$

where $V_x$, $V_y$ are the velocity components of the pixel at location $x$ and $y$ in image $I$. As the major moving object in the field of view, the optical flow provides essential information on the location of LV, as well as its mode of motion, as illustrated in Fig.1, in which the background is clearly suppressed.

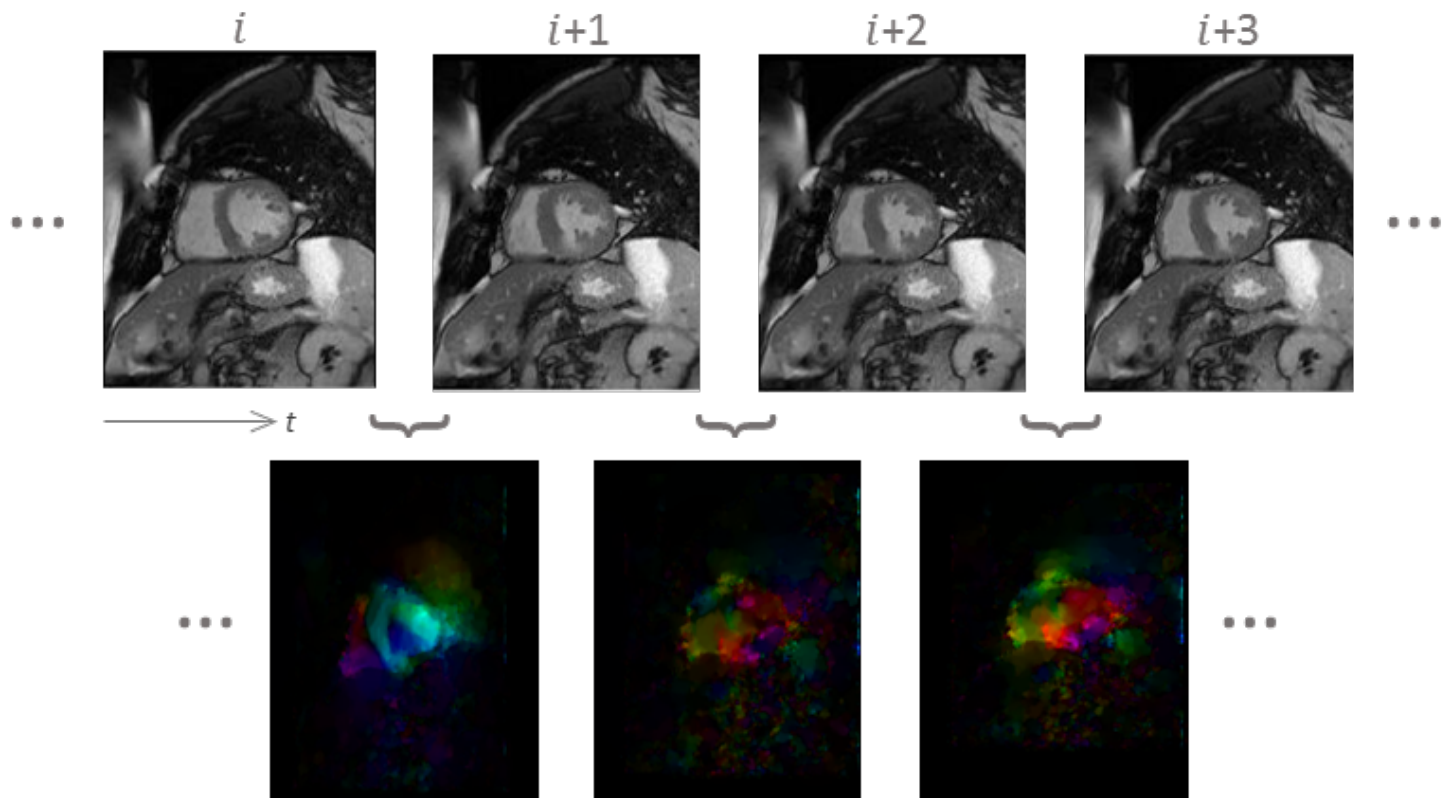

**Fig.1.** Illustration of optical flow in cine MRI between temporal frames. The flow field (lower panel) reflects the local displacement between two frames (upper panel).

## 2.2 Optical Flow Feature Aggregation

We propose to integrate the optical flow into the feature maps, which are extracted by convolutional kernels:

$$m_{j\to i} = I(m_i, O_{j\to i}) \tag{3}$$

where $I(\cdot)$ is the bilinear interpolation function as is often used as a warp function in computer vision for motion compensation [5], $m_i$ represents the feature maps of frame $i$, $O_{j\to i}$ is the optical flow field from frame $j$ to frame $i$, and $m_{j\to i}$ represents the motion-compensated feature maps.

We further aggregated the optical flow information over a longer time span in the cardiac cycle. The aggregated feature map is defined as follows:

$$\bar{m}_i = \sum_{j=i-k}^{j=i+k} w_{j\to i}\, m_{j\to i} \tag{4}$$

where $k$ denotes the number of temporal frames before and after the target frame. Larger $k$ indicates higher capability to follow temporal movement but heavier computation load. We used $k = 2$ as an empirical choice to balance computation load and capture range. The weight map $w_{j\to i}$ measures the cosine similarity between feature maps $m_j$ and $m_i$ at all $x$ and y locations, defined as:

$$w_{j\to i} = \frac{m_j \cdot m_i}{|m_j||m_i|} \tag{5}$$

The feature map $m_i$ and $m_j$ contain all channels of features extracted by convolutional kernels (Fig. 2), which represent low-level information of the input image, such as location, intensity, and edge. Computed over all channels, $w_{j\to i}$ describes local similarity between two temporally neighboring frames. By introducing the weighted feature map, we assign higher weights on locations with little temporal movement for coherent segmentation, while lower weights on locations with larger movement to allow changes.

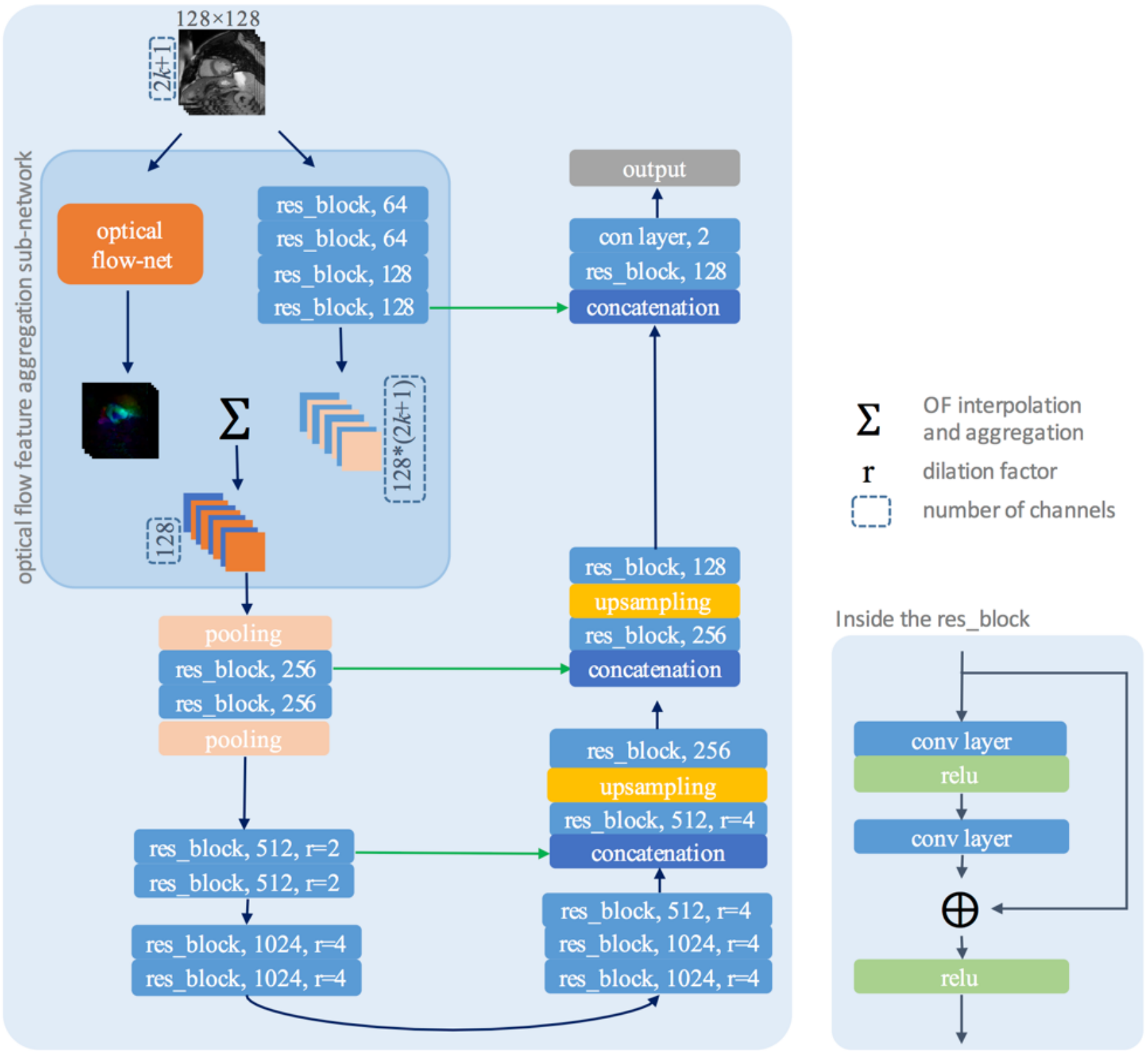

**Fig.2.** The proposed OF-net, including three new characteristics: (1) the optical flow feature aggregation sub-network, (2) res-block, and (3) dilated convolution.

### 2.3 Optical Flow Net (OF-net)

The proposed optical flow feature aggregation is integrated into the U-net architect, which we name as optical flow net (OF-net). The OF-net consists of the following new characteristics compared to the original U-net:

**Optical Flow Feature Aggregation Sub-network:** The first part of the contracting path is made of a sub-network of optical flow feature aggregation described in Section 2.1 and 2.2. With this sub-network embedded, the segmentation of an individual frame takes into consideration information from neighboring frames, both before and after it, and the aggregation acts as a "memory" as well as a prediction. The aggregated feature maps are then fed into the subsequent path, as shown in Fig 2.

**Dilated Convolution:** The max-pooling operation reduces the image size to enlarge the receptive field, causing loss of resolution. Unlike in the classification problem, resolution can be important for segmentation performance. To improve the LV segmentation accuracy, we propose to use dilated convolution [6] to replace part of the max-pooling operation. As illustrated in Fig. 3, dilated convolution enlarges the receptive field by increasing the size of convolution kernels. We replaced max-pooling with dilated convolution in 8 deep layers as shown in Fig. 2.

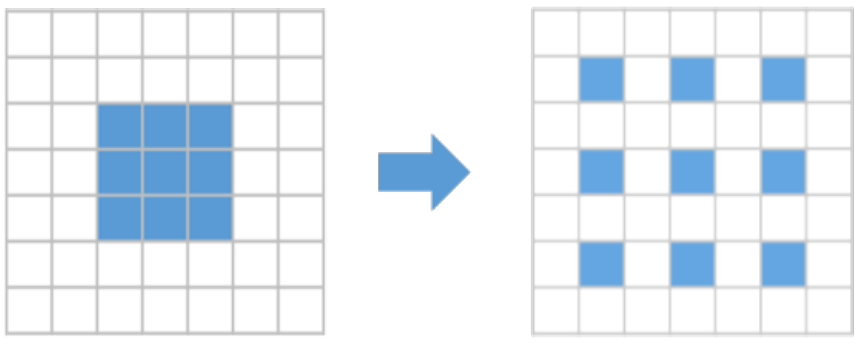

**Fig.3.** Dilated convolution by a factor of 2. Left: the normal convolutional kernel, right: the dilated convolution kernel, which expands the receptive field by a factor of 2 without adding more parameters. Blue indicates active parameters of the kernel while white are inactivated, i.e. set to zero.

**Res-block:** To mitigate the vanishing gradient problem in deep CNNs, all blocks in the U-net (i.e. a convolutional layer, a batch normalization layer, and a ReLU unit) were updated to res-block [7], as illustrated in Fig.2.

The proposed OF-net preserves the U-shape architecture, and its training can be performed the same way as U-net without need of joint-training, as optical flow between MRI frames only need to be computed once. Simplified algorithm is summarized in Algorithm 1. $N_{feature}$, $N_{segment}$ are sub-networks of feature extractor and segmentation, respectively. $P(\cdot)$ denotes computation of optical flow.

**Algorithm 1:** the proposed OF-net.

**input:** $n$ cine MR frames $\{f_i\}$, aggregation parameter $k$

**for** $j = 1{:}n$ **do**

  $m_j = N_{feature}\,(f_j)$

**end for**

**for** $i = 1{:}n$ **do**

  **for** $j = max(1, i-k){:}\,min(n, i+k)$ **do**

    $O_{j\to i} = P(f_j, f_i)$

    $m_{j\to i} = I(m_j, O_{j\to i})$

    $m^n_{j\to i} = softmax\,(m_{j\to i})$

    $w_{j\to i} = \frac{m^n_j \cdot m^n_i}{|m^n_j||m^n_i|}$

  **end for**

  $\overline{m}_i = \sum_{j=max\,(1,i-k)}^{j=min\,(n,i+k)} w_{j\to i}\, m_{j\to i}$

  $p_i = N_{segment}\,(\overline{m}_i)$

**end for**

**output:** predicted segmentation $\{p_i\}$

# 3 Experiments and Results

## 3.1 Data and Ground Truth

Experiments were performed on the short-axis steady-state free precession (SSFP) cine MR images of 100 patients with coronary artery disease and prior myocardial infarction from the Cardiac Atlas database [8]. A large variability exists in the dataset: the MRI scanner systems included GE Medical Systems (Signa 1.5T), Philips Medical Systems (Achieva 1.5T, 3.0T, and Intera 1.5T), and Siemens (Avanto 1.5T, Espree 1.5T and Symphony 1.5T); image size varied from 138×192 to 512×512 pixels; and the number of frames per cardiac cycle ranged from 19 to 30.

Ground truth annotations of the LV myocardium and blood pool in every image were a consensus result of various raters including two fully-automated raters and three semi-automated raters demanding initial manual input. We randomly selected 66 subjects out of 100 for training (12,720 images) and the rest for testing (6,646 images). All cine MR and label images were cropped at the center to a size of 128×128. To suppress the variability in intensity range, each cine scan was normalized to a uniform signal intensity range of [0, 255]. Data augmentation was performed by random rotation within [-30°, 30°], resulting in 50,880 training images.

## 3.2 Network Parameters and Performance Evaluation

We used stochastic gradient descent optimization with an exponentially-decaying learning rate of $10^{-4}$ and a mini-batch size of 10. The number of epochs was 30. Using the same training parameters, 3 CNNs were trained: (1) the original U-net, (2) the OF-net with max-pooling, (3) the OF-net with dilated convolution. The performance of LV segmentation was evaluated in terms of Dice overlap index and average perpendicular distance (APD) between the ground truth and CNN segmentation results. Since LV segmentation is known to have different degree of difficulty at apex, middle, and base, we evaluated the performance in the three segments separately.

## 3.3 Results

The Dice and APD of the three CNNs are reported in Table 1. It can be seen that the proposed OF-net outperformed the original U-net at all segments of LV ($p<0.001$), and with the dilated convolution introduced, the performance is further enhanced ($p<0.001$).

Some examples of the LV segmentation results at apex, middle, and base of LV are shown in Fig.4. It can be observed from (a)-(c) that the proposed method is able to detect a very small myocardium ring at the apex which may be missed by the original U-net. From (g)-(i) it is seen that the OF-net eliminates localization failure at the base. In the middle slices (d)-(f), the OF-net also produced smoother outcome than the original U-net which processes each slice individually. The effect of integrating temporal information is better illustrated in Fig.5, in which we plotted the myocardium (upper panel) and blood pool (lower panel) area, as determined by the resulting endocardial and epicardial contours, against frame index in a cardiac cycle. It can be observed that the results produced by OF-net is smoother and closer to the ground truth than those produced by U-net, showing improved temporal coherence of segmentation.

**Table 1.** Comparison of performance of the three CNNs: (1) the original U-net, (2) the OF-net with max-pooling, (3) the OF-net with dilated convolution. Performance is differentiated at apex, middle, and base of LV. Paired t-test is done comparing (2) and (1), (3) and (2).

| | Apex | | | Middle | | | Base | | |
|---|---|---|---|---|---|---|---|---|---|
| | Dice (%) | APD (pixel) | p value | Dice (%) | APD (pixel) | p value | Dice (%) | APD (pixel) | p value |
| **U-net** | 73.3±4.3 | 1.67±0.25 | | 91.2±4.0 | 1.21±0.16 | | 82.4±4.6 | 1.41±0.19 | |
| **OF-net(max-pooling)** | 81.9±3.2 | 1.19±0.18 | <0.0001 | 92.3±3.6 | 0.95±0.11 | <0.0001 | 86.3±2.9 | 0.99±0.14 | <0.0001 |
| **OF-net(dilated conv)** | **84.5±3.7** | **1.04±0.11** | **<0.0001** | **94.8±3.2** | **0.90±0.09** | **<0.0001** | **89.3±2.5** | **0.94±0.12** | **<0.0001** |

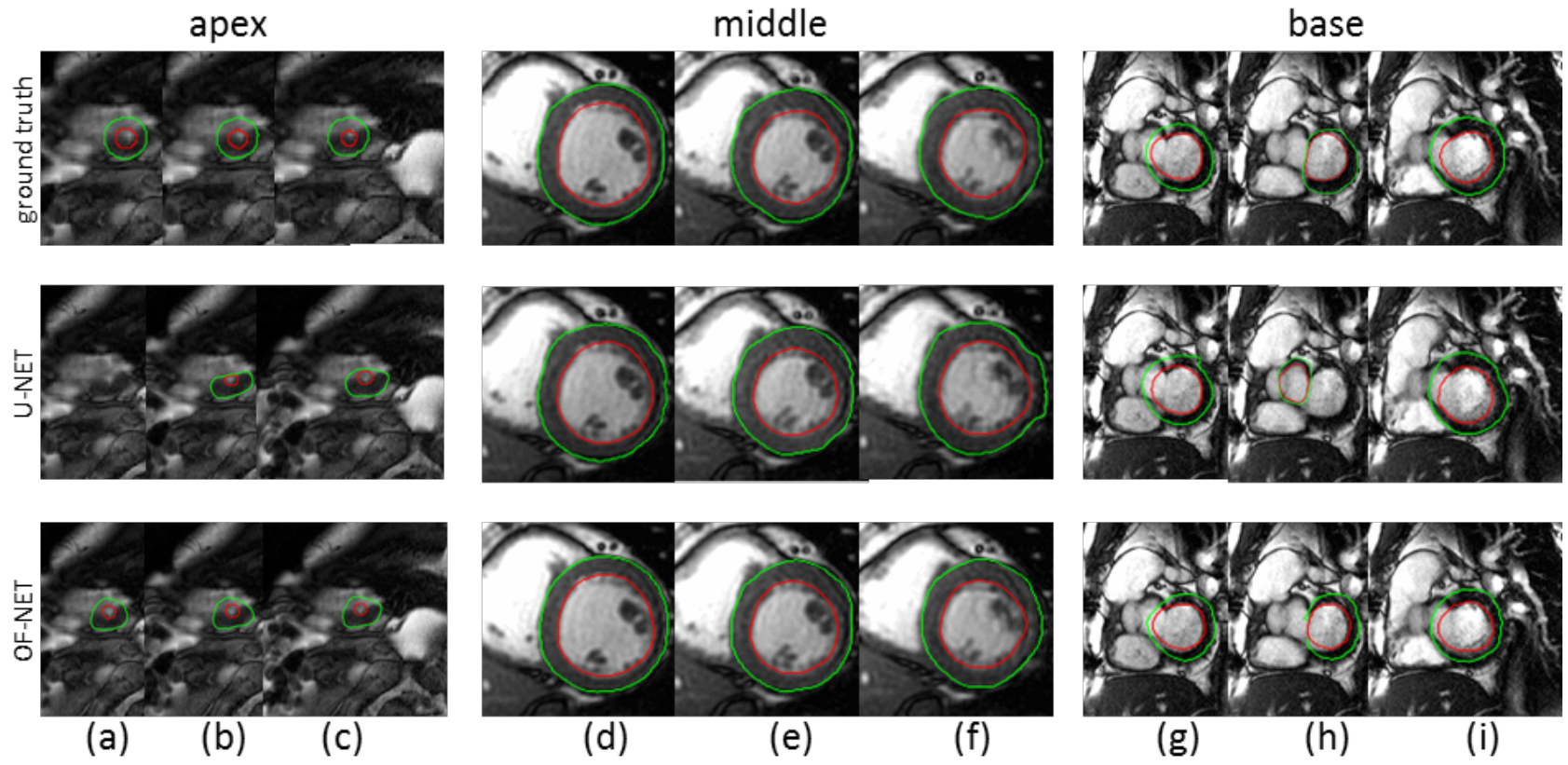

**Fig.4**. Examples of temporal frames at different locations: apex (a)-(c), middle (d)-(f), and base (g)-(i). From top to bottom, the contours are delineated from the ground truth, U-net, and the proposed OF-net, respectively.

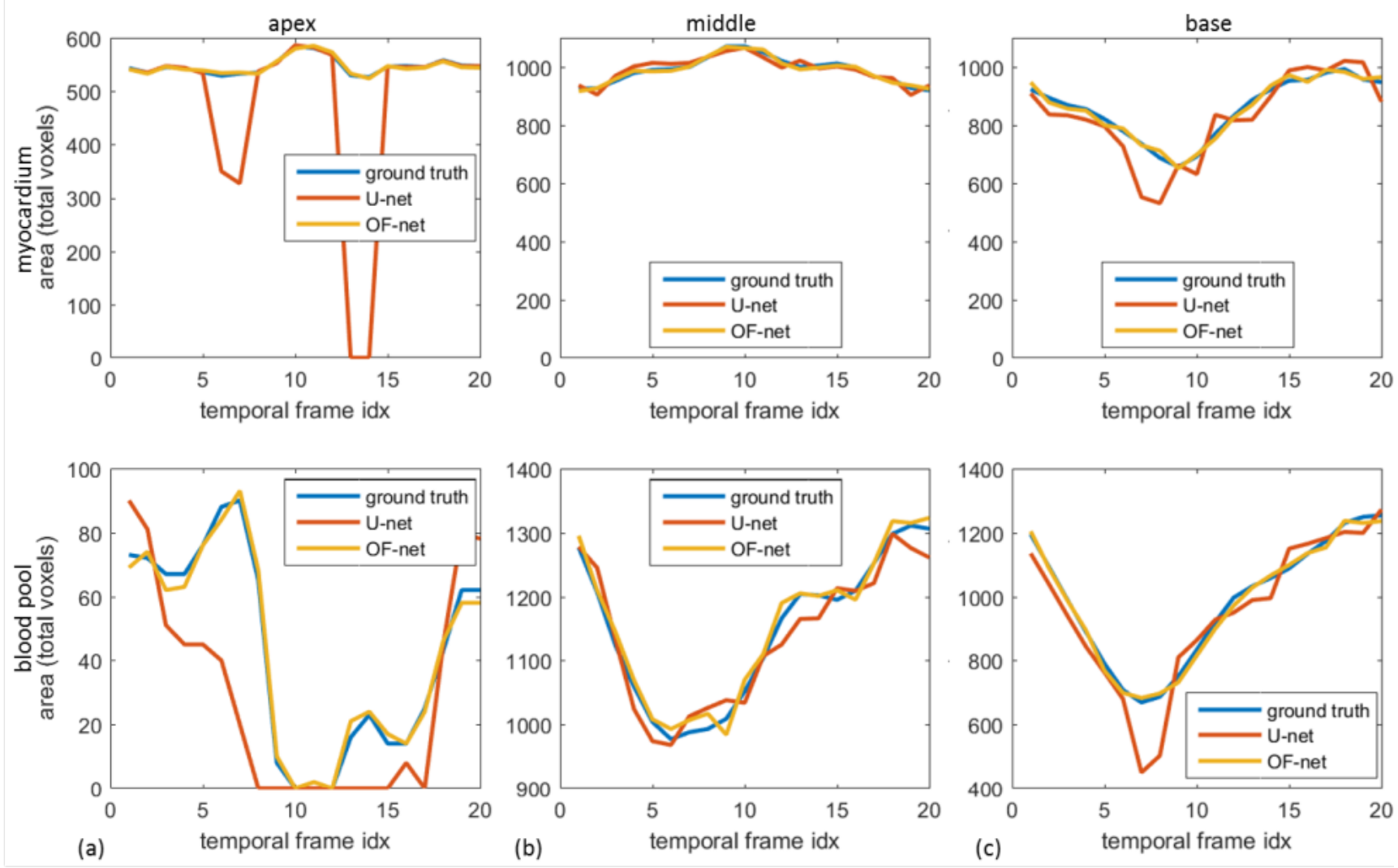

**Fig.5**. Examples of myocardium (upper) and blood pool (lower) area in a cardiac cycle, estimated from the ground truth, U-net, and OF-net, at apex, middle, and base.

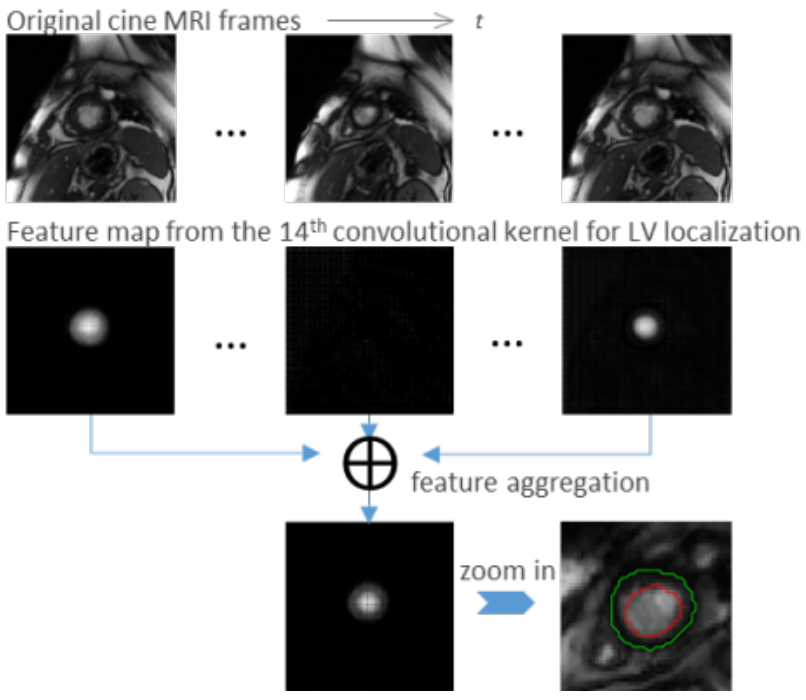

**Fig.6.** Effect of feature aggregation. In the middle frame, the feature map related to "LV location" did not activate. The proposed feature aggregation could retrieve the location of LV based on temporally neighboring slices.

In Fig.6, we illustrate the mechanism how aggregated feature map can help preserve the temporal coherence: the 14$^{th}$ channel in the sub-network is a localizer of LV. While localization of LV in one frame can be missed, the aggregated information from neighboring frames can correct for it and lead to coherent segmentation.

## 4 Conclusion

We have proposed an OF-net for fully automated segmentation of LV from cine MRI. The network integrates temporal information to imitate the expert way of viewing cine. Evaluated on the Cardiac Atlas database, the method outperformed the original U-net, producing more accurate and temporally-coherent LV segmentation.

### References

[1] A. de Roos and C. B. Higgins, "Cardiac Radiology: Centenary Review," *Radiology*, vol. 273, no. 2S, pp. S142–S159, Nov. 2014.

[2] P. Peng, K. Lekadir, A. Gooya, L. Shao, S. E. Petersen, and A. F. Frangi, "A review of heart chamber segmentation for structural and functional analysis using cardiac magnetic resonance imaging," *Magma*, vol. 29, pp. 155–195, 2016.

[3] G. Litjens *et al.*, "A survey on deep learning in medical image analysis," *Med. Image Anal.*, vol. 42, pp. 60–88, Dec. 2017.

[4] W. Xue, A. Lum, A. Mercado, M. Landis, J. Warringto, and S. Li, "Full Quantification of Left Ventricle via Deep Multitask Learning Network Respecting Intra- and Inter-Task Relatedness," *In: MICCAI*, vol. 3, pp. 274–283, Sep. 2017.

[5] X. Zhu, Y. Wang, J. Dai, L. Yuan, and Y. Wei, "Flow-Guided Feature Aggregation for Video Object Detection," *In: ICCV,* pp. 408–417, Oct. 2017.

[6] F. Yu and V. Koltun, "Multi-Scale Context Aggregation by Dilated Convolutions," *In: ICLR*, May 2016.

[7] K. He, X. Zhang, S. Ren, and J. Sun, "Deep Residual Learning for Image Recognition," *In: CVPR*, pp. 770–778, Jun. 2016.

[8] C. G. Fonseca *et al.*, "The Cardiac Atlas Project--an imaging database for computational modeling and statistical atlases of the heart," *Bioinformatics*, vol. 27, no. 16, pp. 2288–2295, Aug. 2011.